\pdfoutput=1
\documentclass[10.5pt]{extarticle}

\usepackage[letterpaper,top=1.15in,bottom=1.1in,left=1in,right=1in,headsep=0.3in,footskip=0.45in]{geometry}
\usepackage{iftex}
\ifPDFTeX
  \usepackage[T1]{fontenc}
  \usepackage{XCharter}
  \usepackage[charter,vvarbb,scaled=1.07]{newtxmath}
  \usepackage[scaled=0.82]{DejaVuSansMono}
\else
  \usepackage{XCharter}
  \usepackage[charter,vvarbb,scaled=1.07]{newtxmath}
\fi
\usepackage{microtype}
\usepackage{amsmath}
\usepackage[numbers,sort&compress]{natbib}
\usepackage{booktabs,array,multirow,makecell,longtable}
\usepackage{graphicx}
\usepackage[table]{xcolor}
\usepackage{pifont}
\usepackage{enumitem}
\usepackage{fancyhdr}
\usepackage{titlesec}
\usepackage{caption}
\usepackage{float}
\usepackage{listings}
\usepackage[most]{tcolorbox}
\usepackage[colorlinks=true,linkcolor=nwpurple,citecolor=nwpurple,urlcolor=nwpurple,
  pdftitle={ScientistCrew: A Multi-Agent Crew of AI Scientists That Writes, Checks, and Polishes Your Paper},pdfauthor={Thanh-Huy Nguyen, Ulas Bagci},pdfkeywords={AI scientist, agentic science, multi-agent systems, LLM agents, automated scientific writing, research automation, reproducibility, citation verification, human-in-the-loop}]{hyperref}

\definecolor{nwpurple}{HTML}{4E2A84}
\definecolor{lavender}{HTML}{ECE8F2}
\definecolor{lavedge}{HTML}{A898C6}
\definecolor{ink}{HTML}{1E1A17}
\definecolor{muted}{HTML}{4B443D}
\definecolor{headgray}{HTML}{EDE8DF}
\definecolor{teal}{HTML}{2F5D62}
\definecolor{tealfill}{HTML}{E3ECEB}
\definecolor{rose}{HTML}{8E2332}
\definecolor{rosefill}{HTML}{F4E6E4}
\definecolor{amber}{HTML}{7A5214}
\definecolor{amberfill}{HTML}{F4EBD8}
\definecolor{errorred}{HTML}{A3202B}
\definecolor{codebg}{HTML}{FBF9F5}
\definecolor{darkbar}{HTML}{2B2530}

\newcommand{\tool}{\textsc{ScientistCrew}}
\newcommand{\pa}{\textsc{PaperAnchor}}
\newcommand{\papertitle}{A Multi-Agent Crew of AI Scientists That Writes, Checks, and Polishes Your Paper}
\newcommand{\repourl}{https://github.com/ScientistCrew/scientistcrew}
\newcommand{\code}[1]{\texttt{#1}}

\newcommand{\ours}{\rowcolor{lavender}}
\newcommand{\best}[1]{\textbf{#1}}
\newcommand{\nagents}{30}
\newcommand{\nwild}{85}
\newcommand{\nfalse}{8}
\newcommand{\nsessions}{93}
\newcommand{\ntokens}{2.6M}
\newcommand{\wallmin}{130}
\newcommand{\sessmin}{299}
\newcommand{\npar}{2.3}
\newcommand{\ntagged}{107}
\newcommand{\nprose}{30}
\newcommand{\nproseold}{123}
\newcommand{\nproseoldmax}{19}
\newcommand{\nprosemax}{6}
\newcommand{\ncite}{17}
\newcommand{\nruns}{5}

\titleformat{\section}{\Large\bfseries\color{ink}}{\thesection.}{0.5em}{}
\titleformat{\subsection}{\large\bfseries\color{ink}}{\thesubsection.}{0.5em}{}
\titleformat{\paragraph}[runin]{\bfseries}{}{0em}{}
\titlespacing*{\section}{0pt}{1.6em}{0.6em}
\titlespacing*{\subsection}{0pt}{1.2em}{0.4em}
\titlespacing*{\paragraph}{0pt}{0.7em}{0.7em}
\DeclareCaptionLabelSeparator{bar}{\;\textbar\;}
\setlist[itemize]{leftmargin=1.4em,itemsep=0.25em,topsep=0.3em}
\setlist[enumerate]{leftmargin=1.6em,itemsep=0.25em,topsep=0.3em}

\fancypagestyle{first}{\fancyhf{}\fancyfoot[R]{\small\thepage}}

\newcounter{finding}
\newtcolorbox{finding}{enhanced,breakable=false,colback=lavender,colframe=lavender,boxrule=0pt,arc=2.5pt,
  borderline west={2.6pt}{0pt}{nwpurple},left=9pt,right=8pt,top=5pt,bottom=5pt,before skip=0.8em,after skip=0.8em,
  before upper={\refstepcounter{finding}\textbf{\textcolor{nwpurple}{Finding \thefinding.}}\hspace{0.45em}}}
\newtcolorbox{card}[2][darkbar]{enhanced,colback=white,colframe=#1,colbacktitle=#1,coltitle=white,boxrule=0.6pt,
  arc=2.5pt,left=6pt,right=6pt,top=4pt,bottom=4pt,toptitle=2.5pt,bottomtitle=2.5pt,fonttitle=\small\bfseries,
  fontupper=\small,title={#2},before skip=0.5em,after skip=0.5em}
\newtcolorbox{promptbox}[1]{enhanced,breakable,colback=codebg,colframe=lavedge,colbacktitle=lavender,coltitle=ink,
  boxrule=0.5pt,arc=2pt,left=6pt,right=6pt,top=4pt,bottom=4pt,fonttitle=\small\bfseries,fontupper=\footnotesize,
  title={#1},before skip=0.6em,after skip=0.6em}
\lstdefinestyle{code}{basicstyle=\ttfamily\footnotesize,columns=fullflexible,upquote=true,keepspaces=true,showstringspaces=false,
  breaklines=true,breakatwhitespace=true,backgroundcolor=\color{codebg},frame=single,rulecolor=\color{lavedge},
  framesep=5pt,xleftmargin=6pt,xrightmargin=6pt,aboveskip=0.7em,belowskip=0.6em,
  keywordstyle=\color{nwpurple},stringstyle=\color{teal},commentstyle=\color{muted}\itshape,
  }
\lstdefinestyle{plain}{style=code,frame=none,backgroundcolor={},xleftmargin=0pt,xrightmargin=0pt,aboveskip=0.2em,
  belowskip=0.2em}
\begin{document}
\thispagestyle{first}

\noindent\raisebox{-0.2em}{\includegraphics[height=13pt]{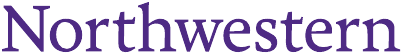}}\hfill{\small\color{muted}November 2026}\par\vspace{5pt}
\noindent\textcolor{nwpurple}{\rule{\linewidth}{0.8pt}}\par\vspace{8pt}
{\raggedright\fontsize{17.5}{22.5}\selectfont\bfseries
\raisebox{-0.14em}{\includegraphics[height=1.12em]{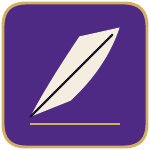}}\hspace{0.18em}\tool{}: A Multi-Agent Crew of AI Scientists\\[1pt]
That Writes, Checks, and Polishes Your Paper\par}
\vspace{8pt}
\noindent\begin{minipage}[c]{0.85\linewidth}
{\raggedright{\bfseries Thanh-Huy Nguyen \quad Ulas Bagci}\par\vspace{2pt}
{\small\color{muted}Northwestern University}\par\vspace{2pt}
{\small\ttfamily\href{\repourl}{\repourl}}\par}
\end{minipage}\hfill
\begin{minipage}[c]{0.14\linewidth}\raggedleft\includegraphics[height=0.5in]{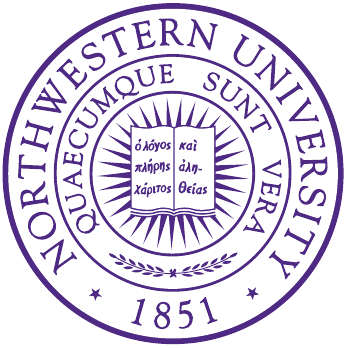}\end{minipage}\par
\vspace{7pt}

\noindent{\bfseries
Agentic science, in which language-model agents act as AI scientists, can now turn a research repository into a
complete paper, but a draft that reads well is not a draft that is right: numbers drift from the data, citations describe a different paper, claims outgrow the experiments,
and figures clip their own labels. We present \tool{}, an open-source multi-agent system of \nagents{} specialist agents for automated scientific writing,
organized the way a lab is. Every hand-off is checked in code before the next agent builds on it: numbers recompute from the raw results, citations are
read against the papers they cite, claims are audited against a ledger of what the data supports, and figures and pages
are inspected at print size. From a small study and two vision studies it wrote CVPR-format papers in which every number recomputes; its
checks stopped \nwild{} errors and flagged every planted miscitation and overclaim. A hand audit of the finished
papers found few wrong numbers but claims wider than the study, which \tool{} now catches by recording what the study
did not test. \emph{A writing agent is only as trustworthy as the hand-offs it checks.}}

\vspace{2pt}
\noindent{\footnotesize\color{muted}\textbf{Keywords:} AI scientist, agentic science, multi-agent systems, LLM agents, scientific writing, reproducibility.}

\vspace{2pt}
\begin{figure}[H]
  \centering
  \includegraphics[width=0.92\linewidth]{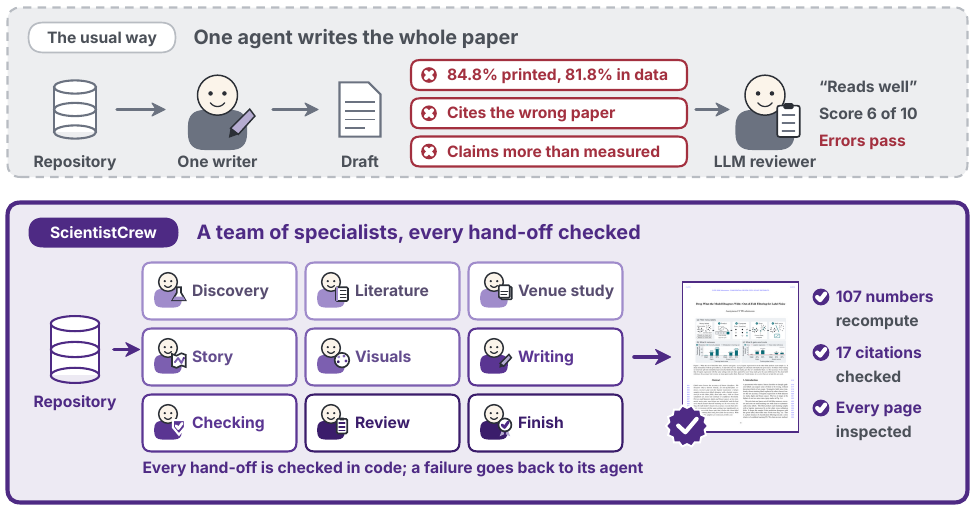}
  \caption{One repository, two ways to write the paper. Top (gray, schematic): the usual way, where one agent drafts and
  revises and its errors (red) read well enough to pass a reviewer. Bottom (purple): \tool{} gives the work to
  specialist teams and checks every hand-off in code, so each error returns to the agent that made it. The purple of
  a team deepens from understanding the repository to review. Right: page~1 of the paper \tool{} wrote from the demo
  study (Section~\ref{sec:results}) and what its checks confirmed.}
  \label{fig:teaser}
\end{figure}

\section{Introduction}
\label{sec:intro}

A researcher finishes a study and asks a coding agent to write the paper. Within an hour the agent has read the result
files, drafted every section, drawn the figures, found references and revised the draft against an LLM reviewer. The
paper reads well. Yet the accuracy in the abstract is not the one in the table, a sentence credits a method to a paper
that proposed something else, a limitation in the brief has become a claim of generality, and a legend sits on top of
the bars it explains. Whole-paper agents already work this way~\cite{lu2024aiscientist,yamada2025aiscientistv2,
schmidgall2025agent,song2026paperorchestra}, and audits of their output report unsupported numbers, failed
integrity checks and hidden methodological flaws~\cite{alfaqeeh2026misattributed,meng2026scientistone,luo2025pitfalls}.

None of these errors is hard to find once someone looks. A number can be recomputed, a citation read against its
abstract, a claim compared with the experiment behind it, a figure inspected at print size. The trouble is that in a
single writer's loop no step looks, and the reviewer at the end of the loop is a poor substitute: LLM reviewers
raise integrity concerns and accept the paper anyway~\cite{badscientist2026}, and human review does not stop such errors either: studies of medicine
and psychology find distorted citations, wrong quotations and misreported statistics in published papers~\cite{greenberg2009how,
jergas2015quotation,nuijten2016statcheck}. Human labs avoid this by dividing the work. The student who ran the
experiments writes the results, a co-author checks the statistics, someone who knows the literature checks the
related work, and the figures go back and forth until they read at a glance. Each hand-off between people is also a
check.

We take this structure literally. \tool{} (Fig.~\ref{fig:teaser}) writes a paper from a research repository with a
team of \nagents{} specialist agents in nine teams: discovery, literature, venue study, story, visuals, writing, checking,
review and finish. Each agent hands its work to the next as a typed file, a setup fact quoted from its source, a recipe that
recomputes a number, a claim with its evidence and strength, a reading card with quotes from an abstract, a figure
spec, a review, and plain Python checks every hand-off before anyone builds on it. A failed check goes back to the
agent that made the error, with the reason, in the same stage. The agents form a dependency graph rather than a line,
so the literature is read while the results are analyzed, and figures, tables, names and references are prepared at
the same time.

Two parts of the team make the paper look like the papers its venue accepts, not merely like a correct report.
Before any writing, \tool{} reads the target venue's own proceedings for the last two years, picks the papers closest
to the work, and turns how they are written into a playbook for framing, Figure~1, contributions, emphasis and
structure. At the end, proofreaders polish each section without rewriting it, and a layout inspector looks at every
rendered page and sends each flaw to its owner, a figure to the agent that drew it and a table to the agent that set
it.

We evaluate \tool{} the way one would evaluate a lab's process: by what it produces, what its checks catch and what
they miss. On a real study it writes a CVPR-format paper in which every printed number recomputes, every citation was
fetched and checked, and every figure and page was inspected, with one storytelling note left open and listed. Across our real runs, its checks stopped \nwild{}
problems, from citation keys that pointed nowhere to tables too wide for their column, and in each case the agent that
caused the problem fixed it. In a planted-defect study, the citation auditor flagged every citation we pointed at the
wrong paper and the claim auditor every claim we widened beyond its evidence, with few false alarms on the sentences we
left alone. Run on an existing paper, the same checks found a misdated reference, two miscitations and a dozen claims
stronger than the evidence. Two people then audited five finished papers by hand, from \tool{} and from a single agent.
Numbers were almost always right. The error that survived in both systems was overreach: a sentence that reads as a
fair summary but names a wider class of methods than the study ran. \tool{}'s auditor had passed these sentences
because each was tied to a ledger claim that the data did support. We therefore made the ledger list what the study
did not test, and checked those limits in code; on the same papers the auditor then caught four of the five
overreaching sentences, against three before. The pattern is the same throughout: \emph{a writing agent is only as
trustworthy as the hand-offs it checks}, and the claim ledger is the hand-off that must say where the evidence ends.

Our contributions are:
\begin{itemize}
  \item \textbf{A checked agent team for paper writing.} We present \tool{}, \nagents{} specialist agents in a
        dependency graph whose every hand-off is a typed file checked in code, with fix rounds that return each error
        to the agent that made it.
  \item \textbf{Venue-aware storytelling.} We show how an agent team can learn a venue's craft from its recent
        papers, and enforce a storyline in code: few numbers in the prose, a framework figure, and emphasis where
        strong papers put it.
  \item \textbf{Evidence of what the checks catch and what they miss.} We measure what the checks caught in real
        runs, how often the agent auditors find planted miscitations and overclaims, and what two people found in
        finished papers, and we show that listing the study's limits in the ledger closes most of the gap they found.
        We release \tool{} as a Claude Code plugin and a command-line tool, with two vision papers as examples.
\end{itemize}

\section{Related work}
\label{sec:related}

\paragraph{Agents that write papers.} End-to-end research agents run experiments and write the paper in one
loop~\cite{lu2024aiscientist,yamada2025aiscientistv2,schmidgall2025agent,weng2024cycleresearcher}, and
data-to-paper systems trace each result back to code~\cite{ifargan2024datatopaper}. Multi-agent writing systems
split the paper among agents, as PaperOrchestra does~\cite{song2026paperorchestra}; its benchmark,
PaperWritingBench, rebuilds the pre-writing materials of 200 published papers and compares systems by reader
preference and citation coverage, but not by whether a printed number matches the materials. Related work on long-form
generation drafts surveys and encyclopedic articles from retrieved sources~\cite{shao2024assisting,
wang2024autosurvey}. These systems show that a full draft is within reach; \tool{} asks what it takes for that draft
to be right, and answers with a check at every hand-off rather than a better writer.

\paragraph{Multi-agent frameworks and self-critique.} General multi-agent frameworks assign roles and let agents
converse~\cite{wu2023autogen,hong2023metagpt,qian2023chatdev}, and self-critique loops let a model improve its own
output~\cite{madaan2023selfrefine,shinn2023reflexion}. Iterated self-critique can also raise its critic's scores while the text gets no better~\cite{pan2024spontaneous}. In \tool{} the roles are fixed by the paper's structure, and every
decision that code can make, whether a number recomputes, whether a quote occurs in an abstract, whether a label
overlaps another, is made by code, so no agent can talk its way past it.

\paragraph{Reviewing and auditing agent-written papers.} LLMs can give useful feedback on
manuscripts~\cite{liang2023can,liu2023reviewergpt}, agent systems simulate or write peer reviews~\cite{jin2024agentreview,goyal2026scholarpeer}, but LLM reviewers miss unsound work and can be gamed by hidden
prompts or by presentation alone~\cite{badscientist2026,sahoo2025injection,nohiddenprompts2026}. Audits of finished papers recompute
numbers or check them against artifacts~\cite{chen2026fabscore,shen2026reagent,meng2026scientistone}; one such study finds that most of the unsupported numbers it labeled in agent-written papers are real values from the wrong experiment, metric or
run~\cite{alfaqeeh2026misattributed}, so matching a value without its identity is not enough. Benchmarks
measure how well models localize errors~\cite{flaws2025}. Our own \pa{}~\cite{paperanchor2026} recomputes each number
at the edit. \tool{} builds on \pa{} for numbers and adds checks at the edit for citations, claims, names, figures and
layout.

\paragraph{Citation and reporting errors.} Citation errors are common in human writing, from distortions that spread
through citation networks~\cite{greenberg2009how} to quotations that misstate their source~\cite{jergas2015quotation},
and language models add fabricated references~\cite{walters2023fabrication,gptzero2026neurips}. Tools check
statistics against their reported tests~\cite{nuijten2016statcheck} and claims against references
semi-automatically~\cite{refverifier2026}. \tool{} fetches every reference from a bibliographic database, so none is
invented, and has an auditor read each citing sentence against the cited abstract before the paper is final.

\paragraph{Figures.} Models can draw scientific figures from text~\cite{belouadi2023automatikz}, and model judges are
now a common way to grade generated output~\cite{zheng2023judging}. \tool{} draws every figure from code that reads
the recomputed values, checks it at print size with a deterministic lint, and only then asks a vision critic whether it
carries its message; the lint decides what code can decide, so the critic is never the only check.

\section{\tool{}}
\label{sec:method}

\tool{} takes a research repository and a short brief from the authors (the idea, the claims they hope to make, where
the results live) and returns a compiled paper in the target venue's template, together with an HTML report that links
every number, citation and figure to its source. Agents do the reading and writing; plain Python decides the order of
work, writes each agent's task, and checks everything that can be checked without a model.

\subsection{Every hand-off is a file, and every file is checked}
\label{sec:handoffs}

An agent in \tool{} never hands free text to the next agent. It writes a file with a fixed form (Table~\ref{tab:handoffs})
and the stage ends with code that checks it. A check that fails returns the problem, in words, to the agent that wrote
the file, which gets up to two fix rounds; the next stage starts only from a file that passes. Agents can run the same
check on their own output with \code{scientistcrew validate}, and most do before they finish.

The checks share one rule: \emph{decide in code whatever code can decide}. A number either recomputes or it does not.
A quote either occurs in the fetched abstract or it does not. A p-value is either below 0.05 or it is not, so a
comparison marked significant with a larger p-value is rejected without asking anyone. Agents are left the judgments
that need reading: whether a cited paper supports a sentence, whether a claim stays within its evidence, whether a
figure carries its message. Even there, code checks the form of the answer, for example that an audit finding quotes
a sentence that is actually in the paper.

\begin{table}[t]
\caption{The hand-offs of \tool{}. Each file is written by the agents on the left and checked in code before the next
stage starts; a failed check returns to the same agents with the reason.}
\label{tab:handoffs}
\centering\small
\begin{tabular}{@{}p{3.3cm}p{3.0cm}p{8.6cm}@{}}
\toprule
Hand-off & Written by & Checked in code \\
\midrule
Setup facts & experimentalist & every fact quotes, word for word, the code or config file it comes from \\
Number recipes & results analyst & every recipe runs on raw data and returns a value; no recipe hard-codes or rounds it \\
Statistical tests & statistician & every comparison names a p-value recipe; its verdict agrees with the value \\
Claim ledger & cartographer & every claim has evidence keys, a scope and a strength; ``strong'' comparisons need a test below 0.05; the ledger lists what the study did not test, and a universal claim counts the settings where it holds \\
Literature shortlist & searcher & every paper came back from a database search \\
Reading cards & readers & every quote occurs in the abstract that code fetched \\
Venue notes, playbook & venue readers, analyst & every quote occurs in the exemplar's PDF; the playbook names the exemplars behind each rule \\
Storyline & three architects & a selection panel compares the pitches in a tournament \\
Outline & outliner & sections name only working recipes and ledger claims; one teaser and one framework figure \\
Names & terminologist & no listed variant appears in the text; abbreviations are defined before use \\
Figures & figure planner, makers & every figure is drawn from recipes and passes the figure lint at print size; a critic may send it back \\
Tables & table setter & tables are rendered from recipe values; a table too wide for its column goes back to the setter \\
Sections & writers, editor & every tagged number recomputes at each edit; every \code{\textbackslash cite} resolves to a fetched entry; no sentence uses a limit term as a finding; the story rules hold \\
Claim audit & claim auditor & every finding quotes a sentence of the paper; every flagged sentence must change \\
Citation audit & citation auditor & every citing sentence is judged; every unsupported citation must change \\
Reviews, meta-review & reviewers, area chair & valid scores and ranked changes; a revision is kept only if every check passes and the score holds \\
Proofread & proofreaders & no section changes by more than a quarter of its words \\
Page inspection & layout inspector & every issue names its owner, who fixes it; the pages are inspected again \\
\bottomrule
\end{tabular}
\end{table}

\subsection{A team in a dependency graph}
\label{sec:graph}

\begin{figure}[t]
  \centering
  \includegraphics[width=\linewidth]{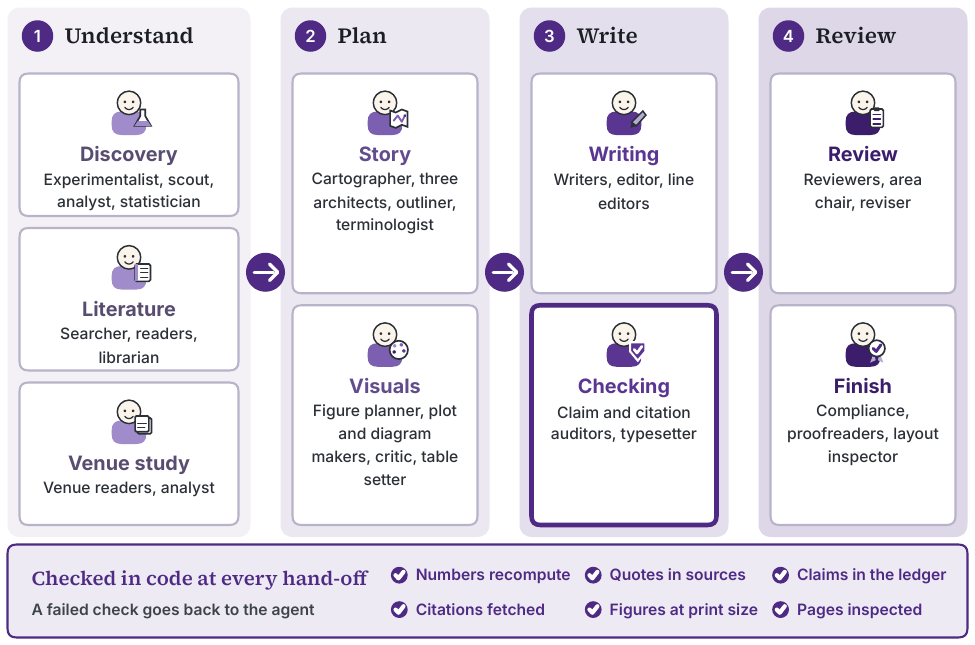}
  \caption{The \tool{} team. Stages start as soon as the stages they need are done, so the teams in a column work at
  the same time. Every arrow is a typed file checked in code (Table~\ref{tab:handoffs}); a failed check returns to the
  agents who wrote the file. The venue study and the literature run while the results are analyzed; proofreading and
  page inspection close the run. The purple deepens from the first phase to the last, and the outlined team audits
  claims and citations.}
  \label{fig:overview}
\end{figure}

The \nagents{} agents form nine teams (Fig.~\ref{fig:overview}). Each stage of work declares the stages it needs, and
an engine starts it as soon as they are done, so the team works the way a lab does: the literature is searched and read
while the results are analyzed and tested, and the names, the bibliography, the figures and the tables are prepared at
the same time once the outline exists. A single cap bounds how many agents run at once, the engine waits out usage
limits and resumes the interrupted sessions, and every accepted stage is committed to a private history, so the
authors can roll the paper back to any stage and run again from there.

\paragraph{Numbers, claims and statistics.} The results analyst writes a recipe for every number the paper may print:
a few lines of Python that read the raw results and return the unrounded value, in the convention of \pa{}. Every
number in the text carries its recipe's key, and the recipe is rerun whenever a writer edits the paper. The
statistician adds a paired test for every comparison the brief or the data suggests, including the ones it expects to
fail. The cartographer then builds the claim ledger, every claim the paper may make with its scope, its evidence and
its strength, and the writers may state only ledger claims, at their scope.

\paragraph{Limits of scope.} A scope written as free text does not stop a writer from widening it. The ledger
therefore also lists the study's limits: the broader classes a reader could generalize to but the study did not test,
each as the short phrase a writer would reach for, what was tested instead, and why it matters (``linear
classifiers'', tested only as logistic regression). Code rejects any sentence that uses a limit term unless it cites
other work, states it as a limit (``only'', ``whether'', ``future work''), or also names what was tested. A ledger
claim that says ``every'', ``all'' or ``consistently'' carries two recipes, the settings where it holds and the
settings it covers, and a strong one must hold in all of them. Every sentence with a universal word needs its own
audit finding. We added these rules after the hand audit in Section~\ref{sec:human}.

\paragraph{Literature and citations.} The searcher maps the field right after the repository is surveyed and may
shortlist only papers that a database search returned. Readers card each paper in parallel from an abstract that code
fetched from OpenAlex, arXiv or Semantic Scholar, with quotes code finds in that abstract. The librarian adds the
official entries, so no reference is ever typed. After editing, the citation auditor reads every citing sentence
against the cited paper's card and abstract and marks it supported, unsupported or unverifiable.

\paragraph{Learning the venue.} When the authors name a top-tier venue, \tool{} reads its open-access proceedings for
the last two years (CVF for CVPR, ICCV and WACV; the NeurIPS proceedings; PMLR for ICML; the MICCAI proceedings; the ACL
Anthology), ranks every title by its relevance to the brief, and downloads the closest papers. Venue readers study how
each is framed, what its Figure~1 shows, how its contributions and emphasis are set, and how its results are told; a
venue analyst turns their notes into a playbook that every writing agent follows. The playbook teaches craft and never
wording. The same step prepares the venue's requirements: the official style files, the page limit, anonymous or
camera-ready mode, required sections such as ACL's Limitations and ICML's Impact Statement, and the NeurIPS checklist,
which a compliance agent answers honestly.

\paragraph{Telling a story.} Three architects pitch competing storylines from the ledger, and a selection panel picks
one in a pairwise tournament. Code then holds the text to a storyline: numbers in the running text are counted per
paragraph, so the tables and figures carry them; the contribution list opens each item with a bold lead-in; related-work
paragraphs open with bold run-in heads; the central claim is in italics once; and one diagram, the framework figure,
shows the main idea with its claim and contributions, and the introduction refers to it. A terminologist fixes the
paper's names before drafting, and code rejects any listed variant.

\paragraph{Figures and pages.} A figure planner gives each figure one message and a chart form; a plot maker or a diagram
artist draws it in code from the recipes, in a theme whose colors stay distinct under color-vision deficiency; a lint
checks text size, overlaps, contrast and colors at print size; and a vision critic sends weak figures back to their
maker. After the reviews, proofreaders polish each section, and a layout inspector reads every rendered page and routes
each flaw to its owner.

\paragraph{Disclosure.} Every paper \tool{} writes carries one visible sentence that cites \tool{} and \pa{}, the checker it is built on, by default:
``Every number in this paper is recomputed from the released results with \pa{} and every reference is checked against
its bibliographic record, by the \tool{} agent pipeline.'' The sentence is true of every paper the tool writes and serves
as the disclosure of AI assistance that venues ask for. The tool prints it at setup and documents it in the README. The
writers may rephrase it or move it to the related work; the checks restore it if it disappears and refuse it if it is
hidden.

\section{How we evaluate \tool{}}
\label{sec:setup}

We ask four questions. Does \tool{} produce a paper that passes its own checks from a real repository, in more than
one field? Do the checks catch the problems agents actually make, and do the agents fix them? When we plant known
errors, how often do the agent auditors find them? And when people read the finished papers, what errors are left?

\paragraph{The study.} The demo study in the \tool{} repository is a small, complete experiment with real results: a
logistic regression, a multilayer perceptron, and logistic regression with out-of-fold label filtering, trained on two
scikit-learn datasets under symmetric label noise of 0\%, 20\% and 40\%, with five seeds each. The repository holds the
training script, its configuration and one file of per-run results, plus a short brief. We chose a small study on
purpose: every claim in the resulting paper can be checked by hand.

\paragraph{The runs.} We ran \tool{} on the study twice with real agents during development, each time with the code as
it then stood, and twice with the full release: the CVPR author kit, anonymous review mode, the \code{standard}
preset (Sonnet agents, an Opus outliner) and a four-page limit, as for a CVPR workshop short paper. The second release
run, made after the figure kit gained its display type, larger labels and aligned arrows, is the case study we report;
all runs, and one audit of an existing paper, contribute to the count of problems the checks caught, which we read
from the runs' logs, one entry per failed check. The runs sat inside our own project folder, so the agents could also
read our general writing guide; the single-agent baselines of our benchmark run outside it.

\paragraph{Two vision studies.} To test the pipeline beyond a small tabular study, we ran two computer vision studies
on GPUs and gave each to \tool{} with a short brief. The first compares heads on frozen DINOv2 features (a linear head,
a convolutional head, and a shallow full-resolution stem we call StemFuse) with a fine-tuned ImageNet U-Net and
feature pyramid network on Oxford-IIIT Pet segmentation, at four label budgets and three seeds, scored by region and
boundary IoU. The second trains one U-Net on COCO crops with clean, averaged, second-shot (Noise2Noise) and blind-spot
targets under synthetic Gaussian noise, counting training data in camera shots. Both studies choose their qualitative
cases by a percentile rule fixed in code. We asked for an eight-page CVPR paper from the first and a four-page CVPR
workshop paper from the second. Both runs used the release code with the \code{standard} preset and ran outside our
project folder.

\paragraph{A hand audit.} To see which errors survive all checks, two people audited five finished papers: two that
\tool{} wrote and three that a single Claude Code agent wrote, given the same repository, brief and venue in one session.
The papers came from three tasks: the demo study, a second small study on class imbalance, and one CVPR 2025 paper
rebuilt from its experiment notes. Each rater received a blinded packet per paper, with the compiled PDF, the
repository's results, and one sheet each for numbers, citing sentences, claim sentences and references. A row is a
number or sentence; the rater labels it correct or names the error. Number sheets list every extracted number except
table cells a script had already matched to the results, of which a quarter were sampled. The raters refined their
sheets with an AI assistant, but every label is theirs. Both rated the CVPR 2025 paper; there, one rater reused the
other's number extraction, so their agreement on that paper is an upper bound. The \tool{} papers in the audit predate
the limit rules of Section~\ref{sec:method}. We later scored two versions of the claim auditor against the people's
labels, running each on the five finished papers.

\paragraph{Planted defects.} To measure the two agent auditors, we took the finished paper of a full run and planted
errors whose ground truth is known by construction. For the citation auditor, we swapped citation keys so that five
citations, in four sentences, point to a paper that does not support them, for example a sentence about robust loss functions now
citing a paper on loss correction. For the claim auditor, we rewrote six claims beyond their evidence: a comparison that
is not significant called significant, a result on one dataset extended to both, a mechanism called measured when it
was not, and a comparison with other methods that was never run. Numbers were left unchanged, so only the auditors
could catch these errors. We reran only the auditor's stage, three times, and counted a planted error as found when the
auditor flagged its sentence. Sentences the original run had judged fine and that we left alone are the controls; a
flag on one of them is a false alarm.

\paragraph{An existing paper.} We ran \code{scientistcrew audit} on the paper that \tool{}~0.1 wrote from the same study,
which predates the claim auditor, the citation auditor and the story rules. The audit copies the paper, compiles it, looks every
reference up in OpenAlex, Crossref and arXiv, and runs the citation auditor, the claim auditor, two reviewers and an
area chair.

\paragraph{Cost.} Every agent session runs through Claude Code on a subscription, and we report cost in tokens and
sessions, read from each session's usage record.

\section{Results}
\label{sec:results}

\subsection{From a repository to a checked CVPR-format paper}
\label{sec:case}

\begin{finding}
Given only the repository and a brief, \tool{} wrote a four-page CVPR-format paper in which every printed number
recomputes, every citation was fetched and read against its source, and every figure and page was inspected, with
independent teams working at the same time throughout.
\end{finding}

\begin{figure}[t]
  \centering
  \includegraphics[width=\linewidth]{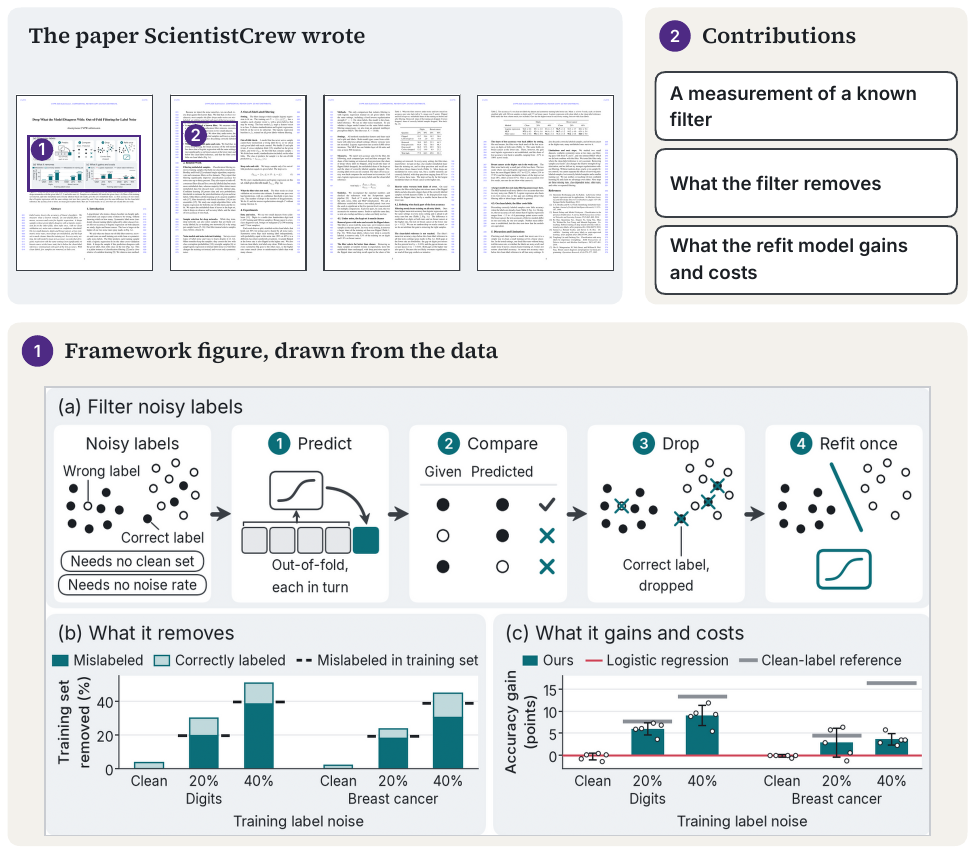}
  \caption{The paper \tool{} wrote from the demo study, in the official CVPR author kit and anonymous review mode. Top
  left: its four pages, with the regions shown below and on the right marked by number. Right: its contribution list,
  each item opening with a bold lead-in (2). Bottom: its Figure~1, the framework figure, at nearly print size (1): the
  four steps of the method, then what the filter removes and what it gains in every setting, weak case included. Its
  tables bold the best result per column.}
  \label{fig:pages}
\end{figure}

\begin{figure}[t]
  \centering
  \includegraphics[width=\linewidth]{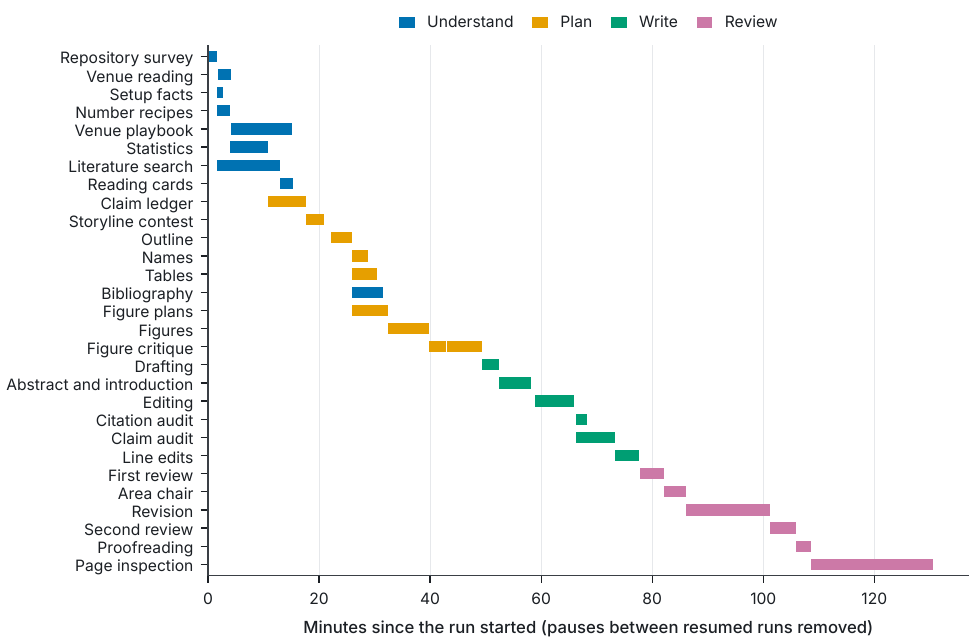}
  \caption{One real run, as a timeline of agent sessions. Each bar is one session, colored by phase; stages in the same
  phase overlap because the engine starts a stage as soon as the stages it needs are done. Repeated bars in a row are
  fix rounds or critique rounds.}
  \label{fig:timeline}
\end{figure}

The release run (Fig.~\ref{fig:pages}) took \nsessions{} agent sessions and \ntokens{} output tokens. Its agents
worked for \sessmin{} minutes in total, but because the engine starts each stage as soon as its inputs exist, the run
took \wallmin{} minutes from start to finish, \npar{} sessions at work at any moment on average (Fig.~\ref{fig:timeline}).
The parallel phases are visible in the timeline: the venue study, the number recipes and the literature search run
together in the first minutes, and the tables, figure plans, names and bibliography start together once the outline is
accepted.

The paper has the shape strong CVPR papers have: a framework figure on page~1 that shows the method in four numbered
steps and, below them, what the filter removes and what it gains; three contributions with bold lead-ins;
related-work paragraphs with run-in heads; results subsections titled by their findings; and an honest limitations
paragraph. All \ntagged{} tagged numbers recompute from the results file, and all \ncite{} cited references were
fetched from bibliographic records. The disclosure sentence sits in the experimental setup, rephrased by the writers
in their own words.

The LLM reviewers scored the paper 4.5 of 10 before and after the revision. Most of their points are about the study,
not the writing: two small datasets, no random-drop or oracle control and no competing noise treatment make thin
evidence for CVPR, and no writer can fix that without new experiments. One point is about the writing: the
introduction does not say what was unknown before the study. One storytelling note stayed open, and the final report
says so: a results paragraph prints six numbers where the rules allow two.

The final check first failed for a reason worth reporting. In the last layout round the typesetter fixed the
typesetting of nine fetched references (dashes in page ranges, an accent in an author's name, a note that makes arXiv
entries print their identifier), and the check, which compared each entry's text with what was fetched, flagged all
nine as edited. None of the edits changed what an entry says. We changed the check to compare a fingerprint of each
entry's content (title, authors, year, venue, pages, identifiers) instead of its text, verified the nine entries
against the versions in the run's history, and the final checks then passed.

\subsection{The same pipeline writes vision papers, and stays inside weak results}
\label{sec:vision}

\begin{finding}
From two GPU studies in computer vision, \tool{} wrote an eight-page CVPR paper and a four-page workshop paper with
teasers, method diagrams, percentile-chosen qualitative results and full result tables, and in both it stated each
result at the strength the data gave it.
\end{finding}

\begin{figure}[t]
  \centering
  \setlength{\fboxsep}{0pt}\setlength{\fboxrule}{0.4pt}
  \fbox{\includegraphics[width=0.31\linewidth]{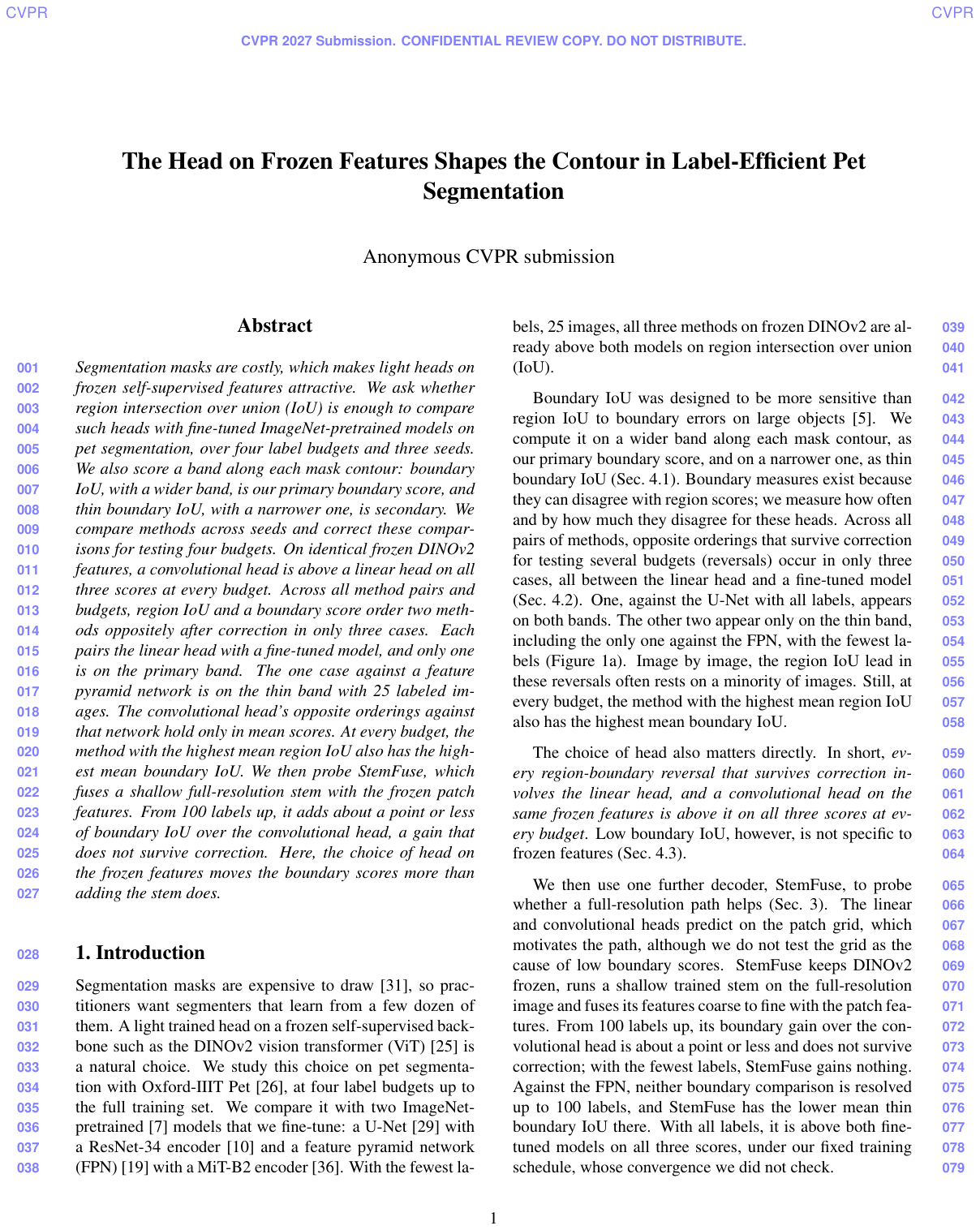}}\hfill
  \fbox{\includegraphics[width=0.31\linewidth]{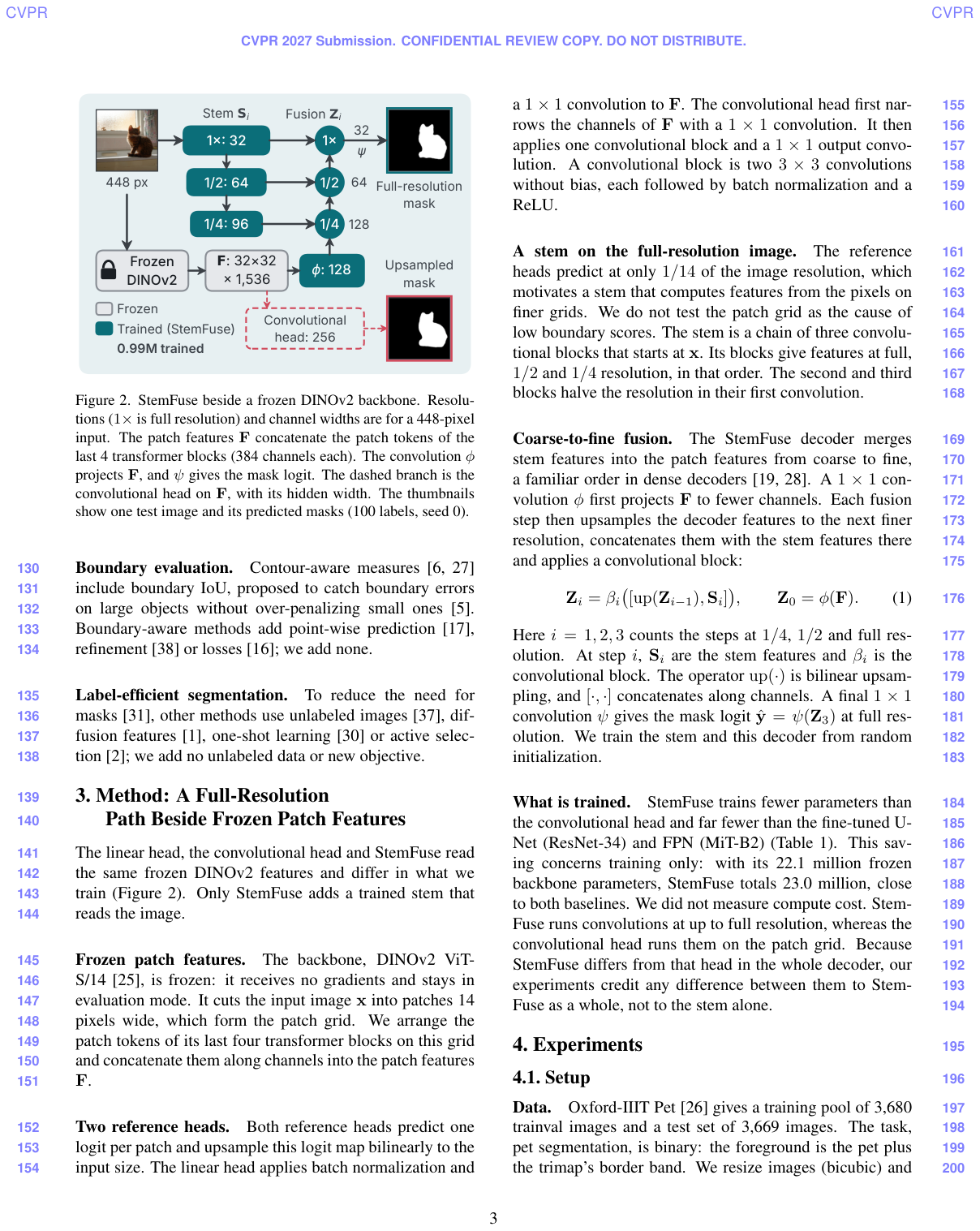}}\hfill
  \fbox{\includegraphics[width=0.31\linewidth]{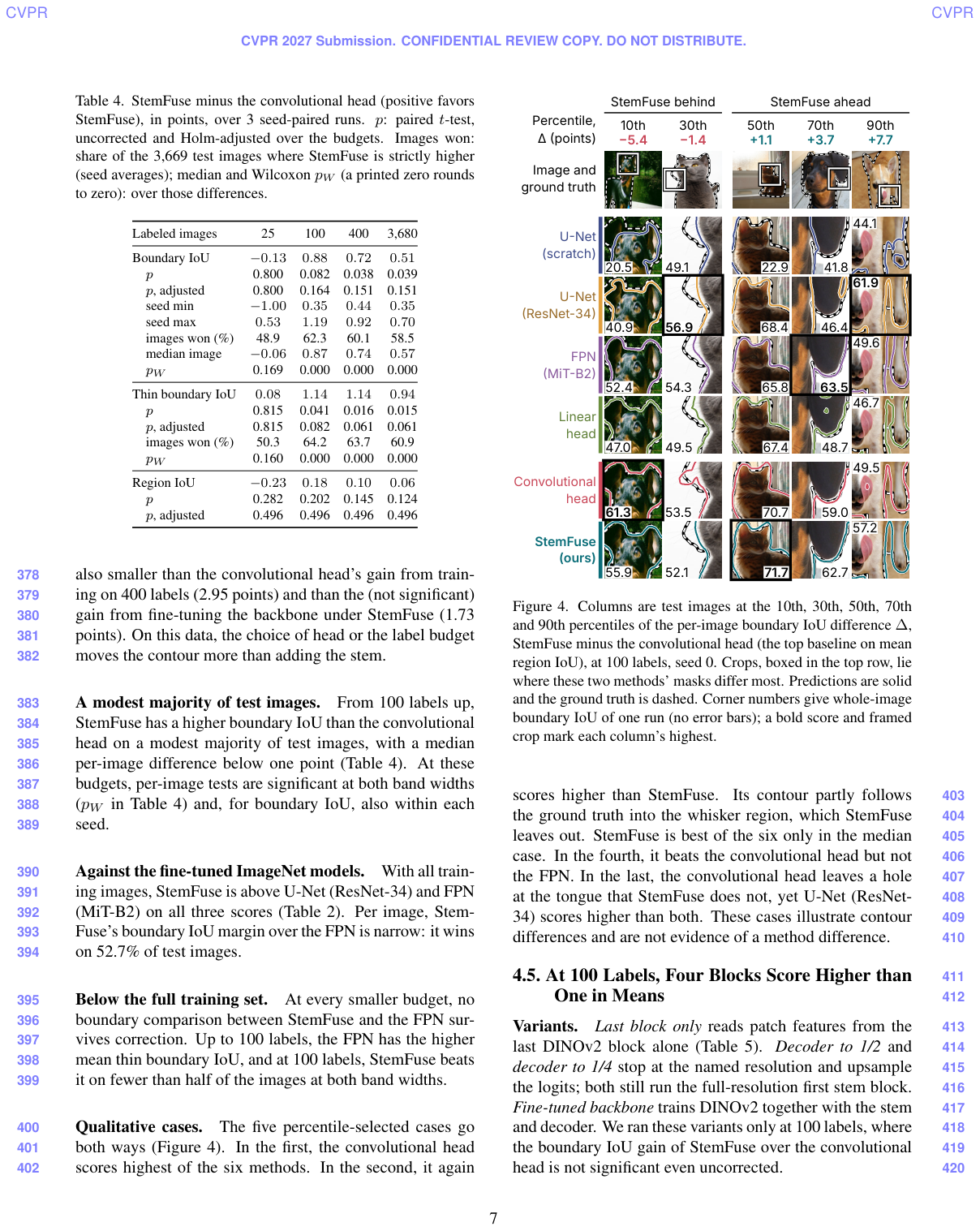}}\\[6pt]
  \fbox{\includegraphics[width=0.31\linewidth]{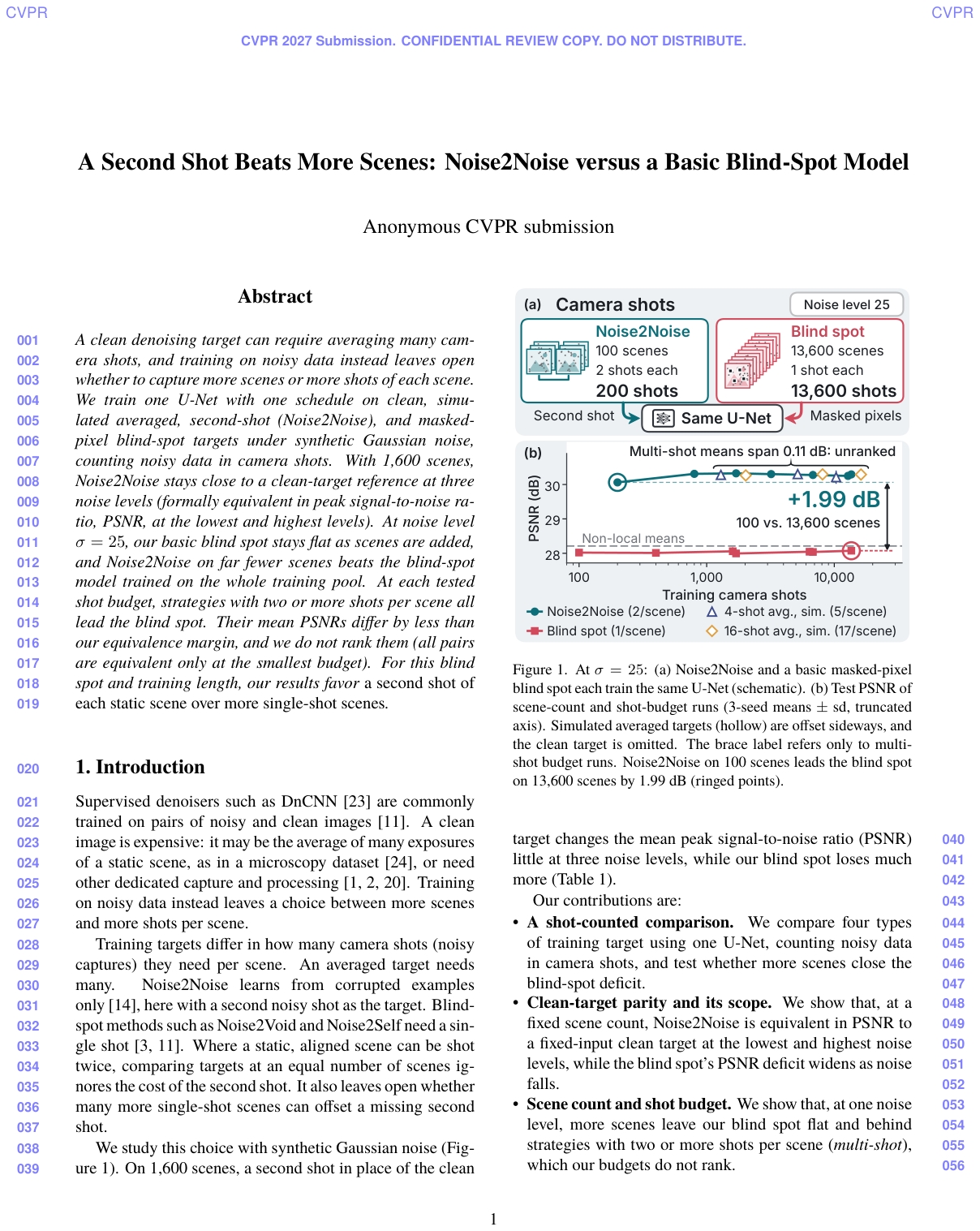}}\hfill
  \fbox{\includegraphics[width=0.31\linewidth]{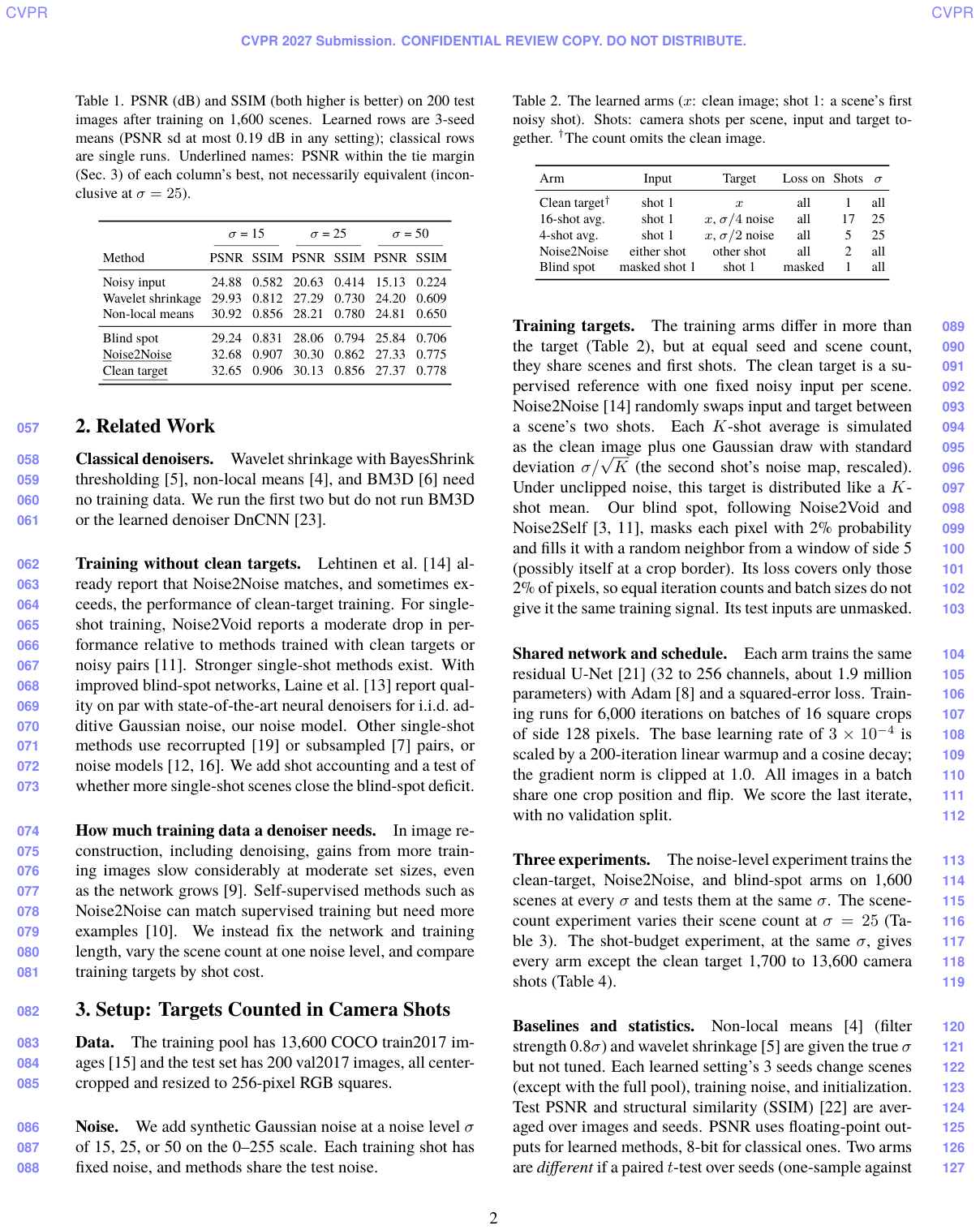}}\hfill
  \fbox{\includegraphics[width=0.31\linewidth]{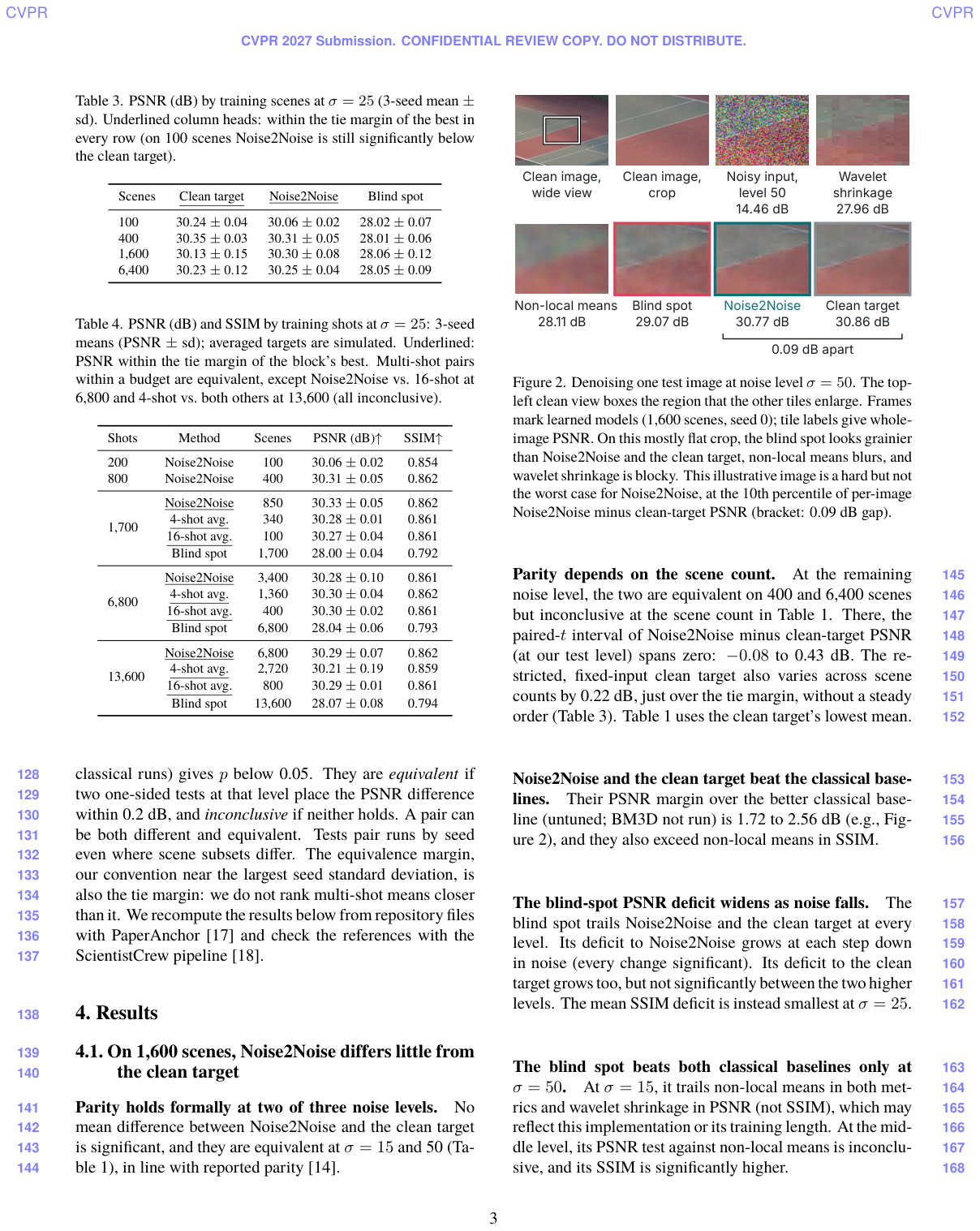}}
  \caption{Two vision papers \tool{} wrote from GPU studies, in the CVPR author kit. Top: pages 1, 3 and 7 of the
  eight-page segmentation paper (teaser with a qualitative inset, the method diagram, and the qualitative grid of test
  images at five percentiles of the per-image difference). Bottom: pages 1 to 3 of the four-page denoising paper
  (teaser, result tables, and a qualitative figure chosen at the 10th percentile). The full papers, their recipes and
  their reviews are in the repository's \code{examples} folder.}
  \label{fig:vision}
\end{figure}

Figure~\ref{fig:vision} shows both papers. The segmentation paper has the parts reviewers look for: a teaser that
pairs the main comparison with a qualitative inset, a method diagram of the stem beside the frozen backbone, label-budget
curves, a qualitative grid chosen by the percentile rule in the study's code, and an ablation table. Its 926 tagged
numbers all recompute from the results. The denoising paper puts the whole argument on page~1: a schematic of two shot
budgets above the curve that compares them.

The two papers differ in what their data allowed. The denoising study has a clear result, a second noisy shot of each
scene beats many more single-shot scenes for the blind-spot model it tests, and the paper tells it plainly; the LLM
reviewers scored it 5 of 10. The segmentation study does not: StemFuse adds less than a point of boundary IoU over the
convolutional head, and the gain does not survive correction for multiple comparisons. The paper says so in its
abstract and its results headings. The reviewers scored it 3.25 of 10, and their objections are fair: the abstract
spends its space on the testing procedure, and the plainest result, that every frozen-feature head beats both
fine-tuned models on region IoU with 25 labeled images, is not in the abstract. The checks kept the paper
honest; they did not make the story clear. Each run took about 4 million output tokens (4.5 million and 3.9 million).

\subsection{The checks stop real errors, and each goes back to the agent that made it}
\label{sec:wild}

\begin{finding}
Across our \nruns{} real runs the checks stopped \nwild{} errors that agents made, from citation keys that pointed nowhere
to tables too wide for their column, and in each case the agent that caused the error fixed it in the same stage.
\end{finding}

\begin{figure}[t]
  \centering
  \includegraphics[width=\linewidth]{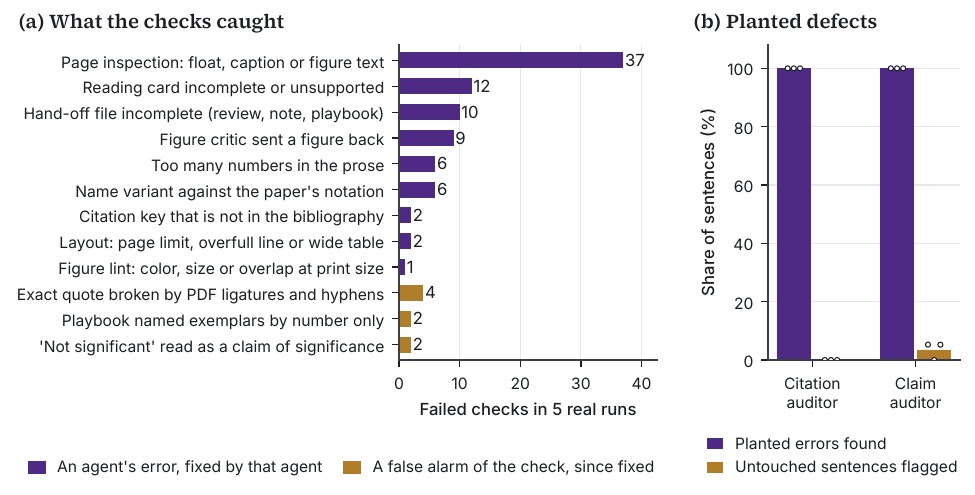}
  \caption{What the checks catch. (a) Every failed check in \nruns{} real runs, by kind; one entry per failed check and
  fix round. Purple: an agent's error, fixed by the agent that made it. Gold: a false alarm of the check itself, each
  since fixed in code. (b) The planted-defect study: the share of planted errors the auditor flagged (purple) and the
  share of untouched sentences it flagged (gold), over three repetitions (dots). Planted citation errors: 5; planted
  claim errors: 6; untouched controls: 6 citations and 19 claims per repetition.}
  \label{fig:checks}
\end{figure}

Figure~\ref{fig:checks}a counts every failed check in the runs' logs. The largest group is layout: the page inspector
found floats printed pages away from their first mention, a caption promising bold entries a table did not have, and
figure text too small to read in print. Next come reading cards that left a field empty or used a relation the
literature map does not know, and hand-off files whose form was incomplete, such as reviews whose weaknesses had no
severity. Figure critics sent figures back for reasons code cannot see, for example a y axis starting at 50 that
exaggerated a drop. The story rules caught paragraphs that listed numbers instead of telling a finding, and the
notation check caught a term that had drifted from ``noise rate'' to ``flip rate'' in four sections at once during a
revision.

The checks were not always right. Three of them raised \nfalse{} false alarms between them: a sentence saying a result
was \emph{not} significant read as a claim of significance, exact quotes from conference papers broken by ligatures and
line-break hyphens in the extracted PDF text, and a playbook that named its sources by number. Each false alarm showed
up in a real run, and each led to a fix of the check, now covered by a test. Because a check's error is visible as a
failed check with a reason, it is found the same way an agent's error is.

\subsection{The auditors find planted miscitations and overclaims}
\label{sec:planted}

\begin{finding}
The citation auditor flagged every citation we pointed at the wrong paper, and the claim auditor every claim we widened
beyond its evidence, in all three repetitions, with no false alarm on the untouched citations and two across the
untouched claims.
\end{finding}

The planted errors (Table~\ref{tab:plants}) are the kind a reader trusts: a sentence about robust loss functions that
cites a paper on loss correction, a mechanism the paper never measured described as confirmed, a result on one dataset
extended to both, a comparison with other methods that was never run. Numbers were left unchanged, so no deterministic
check could catch them. The citation auditor's corrections say what the cited paper actually does and, in most cases, which paper should be
cited instead, for example that ``Patrini et al.\ propose loss correction using the transition matrix, not a robust
loss''. The two flags on untouched claims were
both mechanism sentences, such as ``a logistic regression that has not seen a sample, and disagrees with its label, is
usually right'', which the paper argues but does not measure; a careful co-author might flag them too
(Fig.~\ref{fig:checks}b).

\subsection{People find overreach, not wrong numbers}
\label{sec:human}

\begin{finding}
In five finished papers audited by hand, numbers were almost always right; the error that survived was a claim
stated for a wider class of methods or settings than the study ran, and it survived in \tool{}'s papers as much as in
a single agent's.
\end{finding}

The raters checked 618 numbers and found one wrong: a p-value in a single agent's paper that the results do not
reproduce (Table~\ref{tab:human}). They found one citation that only partly supports its sentence, and no faulty reference. What they found most was
overreach, one or two sentences per paper that read as a fair summary but claim more than was tested. The demo paper
by \tool{} opens its abstract with ``Label noise lowers the accuracy of linear classifiers'', and the single agent's
paper on the same study says ``Label noise hurts simple classifiers''; both studies ran one linear model, logistic
regression. Another paper stated an average as a rule (``F1 does not improve'') while one setting in its own table went the
other way. The one claim the raters disagree on is an ablation that changed several parts at once but credited one. On the two tasks with papers from
both systems, \tool{} made no number error and the single agent one, but the totals are equal at 1.5 errors per paper,
at about ninety times the output tokens. Four papers are too few to compare the systems; they are enough to show
where the remaining errors are.

The audit also explains why \tool{}'s auditor passed these sentences. In the demo run it marked ``Wrong training
labels can cost a linear classifier accuracy'' as supported and linked it to the ledger claim about logistic
regression. Once a sentence was tied to a claim the ledger supports, the auditor accepted its wider wording. The
same auditor, run on the finished paper without a ledger, flags it. \emph{An auditor that checks sentences against
claims inherits the blind spots of the claims.}

\begin{table}[t]
\caption{Errors people found in five finished papers. Rows checked: numbers, citing sentences, claim sentences and
references on each rater's sheet. \tool{} papers predate the limit rules. One claim on the CVPR 2025 paper, on which
the raters disagree, is not counted. Output tokens: the whole writing run.}
\label{tab:human}
\centering\small
\begin{tabular}{@{}llrrrrrrrrr@{}}
\toprule
& & \multicolumn{2}{c}{Numbers} & \multicolumn{2}{c}{Citations} & \multicolumn{2}{c}{Claims} & \multicolumn{2}{c}{References} & Output \\
\cmidrule(lr){3-4}\cmidrule(lr){5-6}\cmidrule(lr){7-8}\cmidrule(lr){9-10}
System & Task & rows & errors & rows & errors & rows & errors & rows & errors & tokens \\
\midrule
\ours\tool{} & demo & 71 & 0 & 11 & 0 & 48 & 2 & 16 & 0 & 2.63M \\
\ours\tool{} & imbalance & 189 & 0 & 14 & 0 & 38 & 1 & 16 & 0 & 3.40M \\
Single agent & demo & 85 & 0 & 14 & 0 & 13 & 1 & 8 & 0 & 0.03M \\
Single agent & imbalance & 214 & 1 & 14 & 0 & 51 & 1 & 11 & 0 & 0.03M \\
Single agent & CVPR 2025 notes & 59 & 0 & 29 & 1 & 29 & 0 & 27 & 0 & 0.04M \\
\bottomrule
\end{tabular}
\end{table}

\subsection{Listing what was not tested catches the overreach}
\label{sec:limits}

\begin{finding}
With the study's limits in hand, the claim auditor caught four of the five overreaching sentences people found,
against three before, with no more false alarms; a check in code alone, with no model judgment, caught two.
\end{finding}

We ran the claim auditor of the release and the auditor with the limit rules on the same five papers and scored their
findings against the people's labels (Table~\ref{tab:limits}). The new auditor first writes down what the paper did
not test, then must give a finding for every sentence that uses one of those terms or a universal word. It adds the
``simple classifiers'' sentence that the release auditor passed, and flags fewer sentences that people accept. Both
miss the irreproducible p-value, which is a number error and needs the number check, not a claim auditor. The limit
check in code, given the auditor's list of limits, flags both overreaching sentences of the demo paper and five sentences people accept in
the same paper, such as ``heavier noise worsens both kinds of error'' over two noise rates. Each costs a writer one
edit to fix. We wrote these rules after reading the labels, so these numbers are in-sample: five errors show the
direction, and the next labeled papers will test it.

The audit also tested the checker we use to score papers automatically. On the labeled papers, its first version
flagged 157 of the 528 numbers people called correct, along with the one real error; most false flags were numbers
the answer key had no entry for, such as a maximum over configurations. Sending those to a person instead of calling them wrong, and re-reading every table the first
pass skipped, cut the flags on correct numbers from 157 to 47 of 528 and raised the share of table cells read from 54\%
to 78\%, while the one real error stayed flagged.

\begin{table}[t]
\caption{Claim audits scored against the people's labels on five papers: 5 sentences people call overreach and 173
they accept. A sentence counts as flagged when a finding that quotes it gives any verdict other than ``ok''. Output tokens: all five
audits. The last row runs no model; it uses the limits the new auditor wrote.}
\label{tab:limits}
\centering\small
\begin{tabular}{@{}lrrr@{}}
\toprule
Claim check & Overreach caught (of 5) & Accepted sentences flagged (of 173) & Output tokens \\
\midrule
Release auditor & 3 & 24 & 256k \\
\ours Auditor with limit rules & \best{4} & 21 & 415k \\
Limit check in code only & 2 & \best{12} & 0 \\
\bottomrule
\end{tabular}
\end{table}

\subsection{The story rules move the numbers into tables}
\label{sec:story}

\begin{finding}
With the story rules, the running text of the paper printed \nprose{} tagged numbers instead of \nproseold{}, and no
paragraph more than \nprosemax{}, while the tables kept every number a reviewer needs.
\end{finding}

The paper \tool{}~0.1 wrote from the same study read like a log: one results paragraph printed \nproseoldmax{} numbers in a
row. The rules in Section~\ref{sec:method} count the numbers per paragraph outside tables and figures, and a writer who
exceeds them must say the rest in words and point to the table. The release paper keeps most of its numbers in two
tables, each with the best result in bold, and its results paragraphs open with the finding.

\subsection{The venue study turns recent CVPR papers into a playbook}
\label{sec:venue}

\begin{finding}
From the open-access CVPR proceedings of the last two years, the venue study picked six papers on noisy labels, and
its analyst turned them into a concrete playbook for this paper; the paper's Figure~1 follows it point by point.
\end{finding}

The ranking used only the brief and the repository: three of the six exemplars (Table~\ref{tab:exemplars}) are
directly about learning with noisy labels, and the other three study noisy labels in multi-label, facial-expression
and medical question-answering settings. The playbook names the exemplars behind each rule, notes where they disagree
with the house rules (some introductions open on the success of deep networks; the house rule, which opens on the
problem, wins), and ends with instructions specific to this paper. One of them shaped the result directly: it asked
for Figure~1 on page~1 with numbered stages, a correctly labeled sample shown being dropped along with the mislabeled
ones, and a small panel of accuracy in every noisy setting, weak case included. The framework figure of the paper does
each of these (Fig.~\ref{fig:pages}).

\begin{card}{From the playbook the venue analyst wrote (excerpt)}
\textbf{Framing.} Name the task or current practice in the first abstract sentence and say what is wrong or missing
by the second (all six; clearest in 03 and 05). Bring in the authors' answer at abstract sentence three (01, 04, 05,
06) or four (02, 03).\par\smallskip
\textbf{For this paper.} Fig.~1 on page 1 (02), with numbered stages (03): noisy training set, out-of-fold
prediction, comparison with the given label, dropped mismatches, refit. Show a correct sample being dropped along with
mislabeled ones. Add a small panel of test accuracy against logistic regression in every noisy setting, weak case
included.
\end{card}

\begin{table}[t]
\caption{The six exemplars the venue study chose for the demo paper, from the CVPR open-access proceedings, ranked by
the relevance of their titles to the brief. No citation key was set, so citations did not enter the ranking.}
\label{tab:exemplars}
\centering\small
\begin{tabular}{@{}rlp{11.2cm}r@{}}\toprule \# & Year & Title & Relevance \\ \midrule
01 & 2025 & Directional Label Diffusion Model for Learning from Noisy Labels & 19.3 \\
02 & 2026 & Revisiting Learning with Noisy Labels: Active Forgetting and Noise Suppression & 17.2 \\
03 & 2026 & Debiased Sample Selection for Learning with Noisy Labels & 13.4 \\
04 & 2025 & Theory-Inspired Deep Multi-View Multi-Label Learning with Incomplete Views and Noisy Labels & 11.5 \\
05 & 2026 & CLEX: Complementary Label Exchange Learning for Noisy Facial Expression Recognition & 11.2 \\
06 & 2025 & DiN: Diffusion Model for Robust Medical VQA with Semantic Noisy Labels & 10.9 \\
\bottomrule\end{tabular}

\end{table}

\subsection{An audit finds what a paper's own pipeline missed}
\label{sec:audit}

\begin{finding}
Run on the paper \tool{}~0.1 wrote, which predates the auditors, the audit found a reference with the wrong year, two
miscitations and claims stronger than the evidence, and its area chair ranked the changes that mattered most.
\end{finding}

All nine references of that paper exist; one, the scikit-learn paper, carried the year 2018 instead of 2011. The
citation auditor found that MentorNet was described as a small-loss selection method, while it learns a curriculum that
weights samples, and that a grouping of the field was credited to a survey that groups it differently. The claim auditor flagged the
paper's description of a known filter as ``a method'', and several sentences that turned a hypothesis into a finding.
The area chair's first required change was the same reframing: present the work as a measurement of a known filter,
not as a new method. None of these errors would have changed a number, so none of them could be caught by the
number check alone.

\section{Using \tool{}}
\label{sec:using}

\tool{} is a Python command-line tool and a Claude Code plugin. At the root of a research repository:

\begin{lstlisting}[style=code]
scientistcrew init --venue cvpr     # title, authors, mode, preset; fetches the official CVPR style files
scientistcrew doctor                # Claude Code, LaTeX, Python packages, the bibliographic databases
scientistcrew run                   # the whole team; paper/main.pdf and .scientistcrew/report.html
scientistcrew watch                 # in a second terminal: stages, agents at work, tokens
\end{lstlisting}

\paragraph{Steering.} \code{scientistcrew run --pause-at story} stops once the storyline is chosen, so the authors can read
or edit the pitch before anyone writes; \code{scientistcrew rollback outline} returns the paper and every note to the
moment the outline was accepted and reruns what came after; \code{scientistcrew history} lists every accepted stage.
Inside an interactive Claude Code session the \code{/scientistcrew:write-paper} skill runs the same stages one at a time,
with the session as the orchestrator.

\paragraph{Presets and venues.} Three presets trade tokens for care: \code{lite} turns off the contests and the
revision rounds and studies three exemplars; \code{standard} runs one revision round with Sonnet agents and an Opus
outliner; \code{max} runs two rounds, more critic rounds and Opus for the story, writing, auditing and review roles.
Seven venues ship with their official style files (NeurIPS, ICLR, ICML, CVPR, ACL, MICCAI and IEEE conferences), and
the venue study covers ten (adding ICCV, WACV, EMNLP and NAACL).

\paragraph{Auditing a paper \tool{} did not write.} \code{scientistcrew audit path/to/paper} copies an existing \LaTeX{}
paper, never changing the original, and runs the checking and review teams on it: the build and its cross-references,
the prose and figure lint, every reference looked up in the databases, every citing sentence read against the cited
abstract, every claim read against the paper's own tables and figures, and venue reviewers with an area chair. The
report lists what to change first.

\paragraph{What the authors still do.} \tool{} writes a first draft that passes its checks; the authors are still the
authors. The HTML report links every number to its recipe and the line that prints it, every citation to its database
record and the sentences that cite it, and every figure to the script that draws it, so a co-author can verify the
draft in the order the checks cannot: the recipes' statements of what they compute, the claims the ledger marks weak,
and the citations the auditor could not verify from an abstract.

\section{Discussion}
\label{sec:discussion}

A paper is a chain of hand-offs: from results to numbers, from numbers to claims, from the literature to citations,
from claims to figures and prose, from prose to the printed page. A single writing agent performs every link of this
chain and checks none of them, so its errors travel to the end, where a reviewer, human or model, is least likely to
catch them. \tool{} checks each link where it is made. The checks that code can make leave nothing to judgment; the
judgments that remain, whether a paper supports a sentence or a claim stays within its evidence, go to auditors whose
findings code still holds to the text. In our runs and our planted-defect study, this structure caught the errors that
single writers are known to make, and sent each one back to the agent that made it.

The hand audit shows where the remaining risk sits. Once numbers are recomputed and citations read, what people still
find is overreach, and it hides in the step that looks safest: a sentence tied to a supported claim inherits that
claim's credibility, whatever its wording. A ledger of what the data supports is not enough; it must also record what
the data does not reach. This is cheap to write down at the moment the evidence is read, and hard to recover later from
fluent prose.

The same structure makes the paper easier to trust after it is written. Because every number has a recipe, every
citation a record and a reading card, and every claim an entry in the ledger, a co-author can verify the draft from the
report instead of re-deriving it, and a reviewer who doubts a number can be shown the line of code that computes it.

\paragraph{Limitations.} Most of our evidence comes from small studies, and the planted-defect set has eleven errors in
one paper. The hand audit covers five papers and five overreaching sentences, and we wrote the limit rules after
reading its labels, so the gain we report is in-sample; it shows the direction, not the size, of the effect. Four papers
are also too few to compare \tool{} with a single agent, and \tool{} spends about ninety times the output tokens. The
limit check catches only the wider terms the ledger names, so a writer can still overreach in words no one listed.
The checks keep a paper honest but do not make its story clear: the segmentation paper stays inside its weak result and
buries its plainest finding. The citation auditor reads abstracts, so a claim that rests on a cited paper's
body is marked unverifiable rather than checked. No check yet compares what the text says about a table or figure with
what the table or figure shows; in an earlier release run a reviewer found such a mismatch that the auditors had passed. The reviewers inside the loop are language models, and their scores
are a signal for keeping or rolling back a revision, not a measure of quality; we have not yet run a study with human
readers. The venue study depends on open proceedings and on bibliographic services that rate-limit unauthenticated
clients, and its value for acceptance is untested. A full run is expensive: the case-study run took \ntokens{}
output tokens for a four-page paper, and the eight-page vision paper about 4.5 million. Finally, the
disclosure citation is enforced only inside \tool{}; the code is open, and a fork can remove it.

\paragraph{What comes next.} The natural next step is a benchmark across many repositories and venues, with human
readers comparing \tool{}'s drafts against those of single-writer agents of equal cost, a hand audit large enough to
test the limit rules out of sample, and a larger planted-defect set for every agent check. The checks themselves suggest a broader view of agent writing: a draft becomes trustworthy not
when a better model writes it, but when every step that can be checked is checked where it happens.

\bibliographystyle{plainnat}
\bibliography{refs}

\clearpage
\appendix
\section{The agents}
\label{app:roles}

Every agent is a Claude Code subagent whose instructions are a Markdown file in the plugin; Table~\ref{tab:roles} lists
them with the description each file gives. The stage that runs an agent writes its task, including the paths of the
files it reads and writes, and the check that follows it.

{\small
\begin{longtable}{@{}p{2.0cm}p{2.6cm}p{11.0cm}@{}}
\caption{The \nagents{} agents of \tool{}, by team, with the description each agent's specification gives (\code{scientistcrew/plugin/agents/*.md}).}\label{tab:roles}\\
\toprule Team & Agent & What it does \\ \midrule \endfirsthead
\toprule Team & Agent & What it does \\ \midrule \endhead
Discovery & experimentalist & Reads the repository's code and configs and records the experimental setup (data, splits, models, training, evaluation, hardware) as facts, each quoted verbatim from the file it comes from. \\
 & results analyst & Results analyst. Writes paper/sources.py, one recipe per number the paper may print, each recomputing that number from the repository's raw data (PaperAnchor convention). \\
 & scout & Surveys a research repository before a paper is written. Finds the method, datasets, baselines, result files and configs, and writes .scientistcrew/inventory.md. \\
 & statistician & Adds a paired statistical test (and the 95\% interval of the difference) for every comparison and trend the paper could claim, as recipes in paper/sources.py, and records each verdict in .scientistcrew/stats.json. \\
Literature & librarian & Builds the bibliography from the readers' cards and the bibliographic databases with `scientistcrew cite`, never typing BibTeX, and writes .scientistcrew/related.md. \\
 & reader & Reads a batch of papers from their fetched abstracts and writes one structured card per paper - what it does, what it finds, how it relates to this work - with quotes code checks against the abstract. \\
 & searcher & Maps the literature a research project sits in - baselines, the methods it builds on, competing approaches, datasets and tools - by searching the bibliographic databases, and writes a shortlist the readers will card. \\
Venue study & venue analyst & Turns the readers' notes on recent papers from the target venue into a concrete writing playbook for this paper (.scientistcrew/venue\_style.md), grounded in the exemplars. \\
 & venue reader & Studies how one recent, relevant paper from the target venue is written - its framing, Figure 1, contribution list, emphasis, structure, how it tells results, its tone - and writes structured notes with exact quotes. \\
Story & architect & Pitches one storyline for the paper under an assigned framing (problem-first, insight-first, capability-first or evidence-first), built only from claims in the ledger. Several architects pitch in parallel and a program committee picks one. \\
 & cartographer & Builds the claim ledger (.scientistcrew/ledger.json), every claim the paper may make with its scope, evidence keys and strength, and maps each claim in the brief to supported, narrowed or unsupported. \\
 & outliner & Plans a paper's storyline, sections, figures and related-work topics as .scientistcrew/outline.json, using only claims the recomputed numbers support. \\
 & terminologist & Fixes the paper's names and notation before drafting - one canonical name per method, dataset, baseline and metric, the forbidden variants, abbreviations and symbols - in .scientistcrew/notation.json, which code then enforces on the text. \\
Visuals & diagram artist & Draws one diagram (the Figure 1 teaser, a method overview or a study design) with the diagram kit (scientistcrew\_diagram) in a flat, illustrative style with short labels, then lints it and looks at it until it is clean. \\
 & figure critic & Looks at one rendered figure as a reader of the printed paper would and decides pass or revise, with concrete fixes. Never edits the figure. \\
 & figure planner & Plans every figure of the paper in detail - the one message, the chart form, the data keys, the layout and a draft caption - in .scientistcrew/figures/plan.json, including the Figure 1 teaser. \\
 & plot maker & Makes one data figure as a Python script with the house style (scientistcrew\_fig), drawing only recipe values, then lints it and looks at it until it is clean. \\
 & table setter & Writes table specs (.scientistcrew/tables/<id>.json) that ScientistCrew renders into booktabs LaTeX from recipe values, so no table holds a typed number. \\
Writing & editor & Turns separately drafted sections into one coherent paper, enforces the page limit, and makes the build, number check, citation check and prose lint pass. \\
 & line editor & Line-edits one section of the paper - fixes the audit's findings, notation and lint problems in that file, then tightens sentences and transitions - without changing what the evidence supports. \\
 & writer & Writes one section of the paper in LaTeX following the outline and style guide, printing every number with its recipe tag and citing only fetched references. \\
Checking & citation auditor & Checks every citation in the drafted paper - each sentence against what the cited paper says - and writes supported, unsupported or unverifiable findings for the line editors. Never edits the paper. \\
 & claim auditor & Audits every claim sentence of the drafted paper against the claim ledger, the numbers and the statistical tests, and writes the findings to .scientistcrew/audit.json for the line editors. Never edits the paper. \\
 & typesetter & Fixes layout problems the build reports - main text over the page limit, overfull lines, badly placed floats - without changing what the paper says. \\
Review & area chair & Area chair. Reads one round of reviews and the paper, separates real problems from noise, and writes a meta-review with a ranked list of required changes the reviser can make without new experiments. Never edits the paper. \\
 & reviewer & Reviews the drafted paper as a critical, fair peer reviewer with an assigned focus and writes a scored review as JSON. Never edits the paper. \\
 & reviser & Revises the paper in response to the consolidated reviews while keeping every check passing; answers points that need new experiments by narrowing claims, never by inventing results. \\
Finish & compliance & Makes the paper meet its venue's formal requirements - required sections such as ACL's Limitations or ICML's Impact Statement, and an honest answer to every NeurIPS checklist question with a pointer to where the paper supports it. \\
 & layout inspector & Looks at every rendered page of the finished paper for layout flaws - overlapping or clipped text, labels off their marks, misaligned panels, badly placed floats, captions, crowded or wasted space - and routes each to its owner. Never edits the paper. \\
 & proofreader & Proofreads one section of the finished paper for grammar, tone, word use, consistency and typos, polishing without rewriting, and logs every change. \\
\bottomrule
\end{longtable}

}

\section{The planted defects}
\label{app:plants}

Table~\ref{tab:plants} lists every planted edit, as applied to the finished paper of a full run before the auditor's
stage was rerun. Each edit applies exactly once, or the study script stops; numbers are never changed.

\begin{table}[H]\caption{The planted defects. Citation plants swap keys so that a sentence cites a paper that does not support it (P4 replaces both keys of one sentence,\allowbreak{} so it plants two miscitations); claim plants widen a claim beyond its evidence. Numbers are never changed.}\label{tab:plants}\centering\scriptsize
\begin{tabular}{@{}p{0.6cm}>{\raggedright\arraybackslash}p{2.0cm}>{\raggedright\arraybackslash}p{6.3cm}>{\raggedright\arraybackslash}p{6.3cm}@{}}\toprule ID & File & Before & After \\ \midrule
P1 & \code{02\_related.tex} & Robust losses change the objective \textbackslash{}citep\{\allowbreak{}zhang2018generalized\} & Robust losses change the objective \textbackslash{}citep\{\allowbreak{}patrini2017making\} \\
P2 & \code{06\_discussion.tex} & memorizes flipped labels \textbackslash{}citep\{\allowbreak{}zhang2016understanding\} & memorizes flipped labels \textbackslash{}citep\{\allowbreak{}brodley1999identifying\} \\
P3 & \code{02\_related.tex} & In the taxonomy of \textbackslash{}citet\{\allowbreak{}frenay2014classification\} this is noise cleansing & In the taxonomy of \textbackslash{}citet\{\allowbreak{}zhang2018generalized\} this is noise cleansing \\
P4 & \code{02\_related.tex} & (a sentence on curriculum and mixture methods) \textbackslash{}citep\{\allowbreak{}jiang2018mentornet,\allowbreak{}li2020dividemix\} & (same sentence) \textbackslash{}citep\{\allowbreak{}patrini2017making,\allowbreak{}zhang2016understanding\} \\
\midrule
C1 & \code{abstract.tex} & and not significant at 20\textbackslash{}\% noise & and significant at every noise rate \\
C2 & \code{abstract.tex} & On digits,\allowbreak{} refitting wins back most of the mean accuracy gap & On both datasets,\allowbreak{} refitting wins back most of the mean accuracy gap \\
C3 & \code{05\_results.tex} & The filter beats it on digits at both noise rates and on breast cancer at 40\textbackslash{}\% & The filter beats it at every noise rate on both datasets \\
C4 & \code{05\_results.tex} & Neither change is significant,\allowbreak{} and both intervals include zero & Both changes are significant improvements,\allowbreak{} and both intervals include zero \\
C5 & \code{05\_results.tex} & so it fits the flipped labels. & so it memorizes the flipped labels,\allowbreak{} which we confirm by measuring its training accuracy. \\
C6 & \code{abstract.tex} & We tuned neither \$C\$ nor \$K\$ and compared with no other noise-robust method,\allowbreak{} so this is a measurement for one setting,\allowbreak{} not a ranking. & Out-of-fold filtering outperforms existing noise-robust methods. \\
\bottomrule\end{tabular}\end{table}

\section{The disclosure sentence and its checks}
\label{app:disclosure}

The sentence \tool{} adds when no sentence of the main text cites it is: ``Every number in this paper is recomputed
from the released results with \pa{} and every reference is checked against its bibliographic record, by the \tool{}
agent pipeline.'' In the case-study run the writers rephrased it as ``We recomputed the reported numbers with
PaperAnchor and checked the references with the ScientistCrew agent pipeline'', placed it in the experimental setup, and it passes. The checks reject a citation in a comment or an \code{\textbackslash iffalse} block, one made invisible by
white or tiny text, one in a file the main file does not input, and a bare citation without a sentence of at least
eight words; they restore the entry if it is missing or edited, and the plugin refuses an agent edit that would delete
the last citation.

\end{document}